%% file: arxiv.tex
\documentclass[10pt,letterpaper]{fairmeta}

\usepackage[T1]{fontenc}
\usepackage{lmodern}
\usepackage{microtype}
\input{preamble}

\usepackage[numbers,sort&compress]{natbib}
\usepackage[font=small,labelsep=period]{caption}
\usepackage[hidelinks,breaklinks]{hyperref}
\usepackage[capitalize]{cleveref}
\makeatletter
\patchcmd{\Gread@pdf}{\Gread@transgrouptrue}{\Gread@transgroupfalse}{}{}
\makeatother
\crefname{section}{Sec.}{Secs.}
\Crefname{section}{Section}{Sections}
\crefname{table}{Tab.}{Tabs.}
\Crefname{table}{Table}{Tables}

\definecolor{wacvblue}{rgb}{0.21,0.49,0.74}

\RenewDocumentEnvironment{abstract}{+b}
  {\gdef\abstractlist{{\color{metafg}#1}}}{}

\AtBeginEnvironment{table}{\let\scriptsize\small}

\title{SixthView: A Benchmark for Hidden-View Reconstruction and 3D Assembly Reasoning}
\author[1,2,*\dagger]{Pooya Khosravi}
\author[1,\dagger]{Jen-Feng Chang}
\author[1,\dagger]{Fnu Pramono}
\affiliation[1]{Meta}
\affiliation[2]{University of California, Irvine}
\contribution[*]{Work done during an internship at Meta}
\contribution[\dagger]{All authors contributed equally.}
\hypersetup{
  pdftitle={SixthView: A Benchmark for Hidden-View Reconstruction and 3D Assembly Reasoning},
  pdfauthor={Pooya Khosravi, Jen-Feng Chang, Fnu Pramono}
}

\begin{document}
\input{sec/0_abstract}
\maketitle
\input{sec/1_intro}
\input{sec/2_related}
\input{sec/3_benchmark}
\input{sec/4_certification}
\input{sec/5_experiments}
\input{sec/6_results}
\input{sec/7_failure}
\input{sec/8_conclusion}

\bibliographystyle{ieeenat_fullname}
\bibliography{main}

\clearpage
\appendix
\phantomsection
\pdfbookmark[0]{Supplementary Material}{supplementary-material}
\begin{center}
  {\Large\bfseries Supplementary Material}
\end{center}
\setcounter{figure}{0}
\setcounter{table}{0}
\setcounter{equation}{0}
\renewcommand{\thefigure}{S\arabic{figure}}
\renewcommand{\thetable}{S\arabic{table}}
\renewcommand{\theequation}{S\arabic{equation}}
\renewcommand{\theHfigure}{S\arabic{figure}}
\renewcommand{\theHtable}{S\arabic{table}}
\renewcommand{\theHequation}{S\arabic{equation}}
\input{sec/supplement}
\input{sec/supp_references}

\end{document}

%% file: preamble.tex
\usepackage{amsmath,amssymb}
\usepackage{booktabs}
\usepackage{multirow}
\usepackage{tabularx}
\usepackage{graphicx}
\usepackage{float}
\usepackage{tikz}
\usepackage{fvextra}
\usepackage{xspace}
\usetikzlibrary{arrows.meta,positioning}

\newcommand{\benchmark}{{\upshape\textsc{SixthView}}\xspace}

%% file: sec/0_abstract.tex
\begin{abstract}
Inferring the unseen side of a 3D assembly requires reasoning about part placement
and occlusion. Different assemblies can match the observed views while producing
different hidden views, leaving the target ambiguous. We introduce \benchmark, a benchmark
in which models predict a sixth orthographic view from five observed views and
information about the component shapes. The
benchmark contains 602 synthetic polycube assemblies with 2--10 parts and requires
a dense grid of part labels for the missing view, allowing both exact-view and
cell-level evaluation. For each item, we exhaustively verify that all legal
assemblies consistent with the exact component geometry and observed label grids
yield the same target view. With these inputs, the task tests symbolic assembly
and coordinate reasoning. The image-only variants additionally require shape
recovery from rendered views, where occlusion can make different 3D shapes look
identical. We evaluate three vision-language models under four input conditions.
Across 552 items with complete evaluations, exact-grid accuracy with images plus
exact text ranges from 33.2\% to 85.4\%. Adding exact text to separate images
improves accuracy by 19.7--63.0 percentage points, whereas adding images to exact
text changes it by only $-0.5$ to $+2.1$ points. Separate image panels outperform
a merged sheet by 6.7--11.1 points. The results show that performance depends
strongly on access to exact symbolic information, with little further improvement
from images once the geometry is specified.
GitHub: \url{https://github.com/facebookresearch/sixthview_benchmark}.
\end{abstract}

%% file: sec/1_intro.tex
\section{Introduction}
\label{sec:intro}

Humans routinely infer three-dimensional structure from a small collection of
views. This ability supports part assembly, mechanical visualization, and
other forms of spatial problem solving
~\cite{shepard1971mental,hegarty2004mechanical}. Current
vision-language models (VLMs) remain brittle when the viewpoint changes or the
evidence spans several images~\cite{fu2024blink,ma20253dsr,wang2026mindcube,zhang2026multiview}.
Most benchmarks expose this brittleness through a multiple-choice relation or a
coarse action. A model can therefore succeed without committing to a complete,
checkable geometric interpretation.

Discrete orthographic view completion offers a focused diagnostic. Given
several projections, the solver must align coordinate frames, associate object parts across views,
reason through occlusion, and determine the first visible part on every output
ray. Predicting a dense label grid requires more explicit geometric commitments
than selecting a relation or an action. Each cell makes a local error observable.
When exact component coordinates are supplied, the task primarily tests symbolic
assembly and coordinate reasoning; it does not require recovering those shapes
from images. We use \emph{reasoning} to describe these required task operations,
not an inference about a model's internal computational process.

This task is meaningful only when the observations determine the requested
view. Orthographic projection discards depth, and distinct solids can share the
same projections, much as a collection of silhouettes specifies only a visual
hull~\cite{laurentini1994visual}. Even known component shapes may admit several
legal placements that agree with every observed surface. If those placements
produce different hidden views, scoring against one annotated grid confounds a
reasoning error with an underdetermined question. Target uniqueness is thus a
prerequisite for valid evaluation and must be established for each generated
item.

Motivated by these requirements, we introduce \benchmark, a validated diagnostic
benchmark for structured 3D spatial reasoning and assembly, expressed as dense
hidden-view prediction. Our work makes three connected
contributions. First, we provide \textbf{a benchmark with complete, localized
predictions}. Its 602 problems span workspaces from $3^3$ to $10^3$ and contain
100 or 101 items per held-out direction. Second, we provide \textbf{a
validated construction procedure}. It generates connected labeled
polycubes, screens them for simple shortcuts, and automatically checks that each
benchmark item's five first-hit label grids and exact labeled piece shapes
determine one target under all proper rotations and in-bounds translations.
This validation scope directly covers the two conditions that provide
exact piece geometry. The two image-only conditions serve as secondary
diagnostics because static piece renders can leave geometry ambiguous.
Third, we present \textbf{an item-matched three-model
evaluation}. Four input conditions and three repeated trials measure sensitivity
to exact geometry and image layout. A sanitized ledger of final parsed grids,
statuses, hashes, and scores supports recomputation without raw rationales or
model-internal reasoning. These synthetic assembly problems provide a focused
spatial diagnostic; transfer to natural images or physical assembly remains
outside the benchmark's scope.

%% file: sec/2_related.tex
\section{Related work}
\label{sec:related}

\paragraph{Spatial reasoning diagnostics.}
CLEVR and GQA established generated scenes and targeted splits as diagnostic
tools~\cite{johnson2017clevr,hudson2019gqa}. BLINK, PureSpace, and psychometric
evaluations expose gaps in perception and spatial
abilities~\cite{fu2024blink,li2026purespace,xu2025psychometric}.
SPHERE organizes human-annotated spatial questions into single-skill,
multi-skill, and reasoning levels; counting uses open-ended numerical answers,
while other tasks use multiple choice~\cite{zhang2025sphere}.
3DSRBench evaluates height, orientation, location, and multi-object relations
with visual questions on natural images, including viewpoint-robustness
subsets~\cite{ma20253dsr}. These benchmarks diagnose broader visual spatial
skills than the discrete assembly operations isolated by \benchmark.

\paragraph{Multiview evidence and answer formats.}
MuirBench evaluates diverse multi-image relations with multiple-choice
questions and pairs standard instances with minimally changed unanswerable
variants~\cite{wang2025muir}. CityCube extends cross-view evaluation to urban
environments, using multiple-choice questions across camera movements and
platforms such as vehicles, drones, and satellites~\cite{xu2026citycube}.
MultiView-Bench targets world-centric integration and asks for an object's
relative position along three global axes as a compact relation
tuple~\cite{zhang2026multiview}. VSI-Bench, MindCube, VIEW2SPACE, and
SpatialMosaic offer further tests of integration across partial
observations~\cite{yang2025vsi,wang2026mindcube,ke2026view2space,lee2026spatialmosaic}.
Thus multiview integration, coordinate reasoning, and answerability are already
established evaluation goals. \benchmark contributes their combination with a
dense label for every target ray and an exhaustive check that all assemblies
consistent with exact geometric evidence agree on that target.

\paragraph{Orthographic geometry and validation.}
SPARE3D, SpatialViz-Bench, TACIT, and 3ViewSense study projection and view
completion~\cite{han2020spare3d,wang2026spatialviz,medeiros2026tacit,zhan20263viewsense},
while CReFT-CAD, Ortho2CAD, MechVQA, and OmniMech target richer engineering
artifacts~\cite{niu2025creft,ortho2cad2026,kou2026mechvqa,lu2026omnimech}.
PolyComp is closest in using polycubes and exhaustive rigid transformations,
but verifies candidate composition rather than whether observations identify
one scored target~\cite{patel2026polycomp}. \benchmark guarantees a unique
held-out target even when several latent assemblies remain feasible. It applies to the
conditions supplying exact component coordinates and observed label grids;
the image-only variants remain uncertified diagnostics. This defines a
complementary benchmark for structured assembly, without claiming a general
measure of visual 3D understanding.

%% file: sec/3_benchmark.tex
\section{Task and benchmark construction}
\label{sec:benchmark}

\begin{figure*}[t]
  \centering
  \begin{tikzpicture}[
    node distance=5mm,
    stage/.style={draw=black!45, rounded corners=2pt, fill=wacvblue!6,
      align=center, text width=0.205\textwidth, minimum height=12mm,
      inner sep=3pt, font=\footnotesize},
    flow/.style={-{Latex[length=1.8mm]}, semithick, draw=black!60}
  ]
    \node[stage] (generate) {\textbf{Generate}\\Connected labeled polycubes\\$3^3$ to $10^3$ workspaces};
    \node[stage, right=of generate] (observe) {\textbf{Render}\\Five projections and separated parts\\One view withheld};
    \node[stage, right=of observe] (certify) {\textbf{Validate}\\One target from exact shapes\\and five label grids};
    \node[stage, right=of certify] (evaluate) {\textbf{Evaluate}\\Two exact-geometry conditions\\Two visual diagnostics};
    \draw[flow] (generate) -- (observe);
    \draw[flow] (observe) -- (certify);
    \draw[flow] (certify) -- (evaluate);
  \end{tikzpicture}

  \vspace{1mm}
  {\footnotesize 602 items \quad $\vert$ \quad 100--101 per held-out direction
  \quad $\vert$ \quad 3 evaluated models \quad $\vert$ \quad dense outputs}

  \vspace{1.5mm}
  \begin{minipage}[t]{0.475\textwidth}
    \centering
    \includegraphics[width=\linewidth]{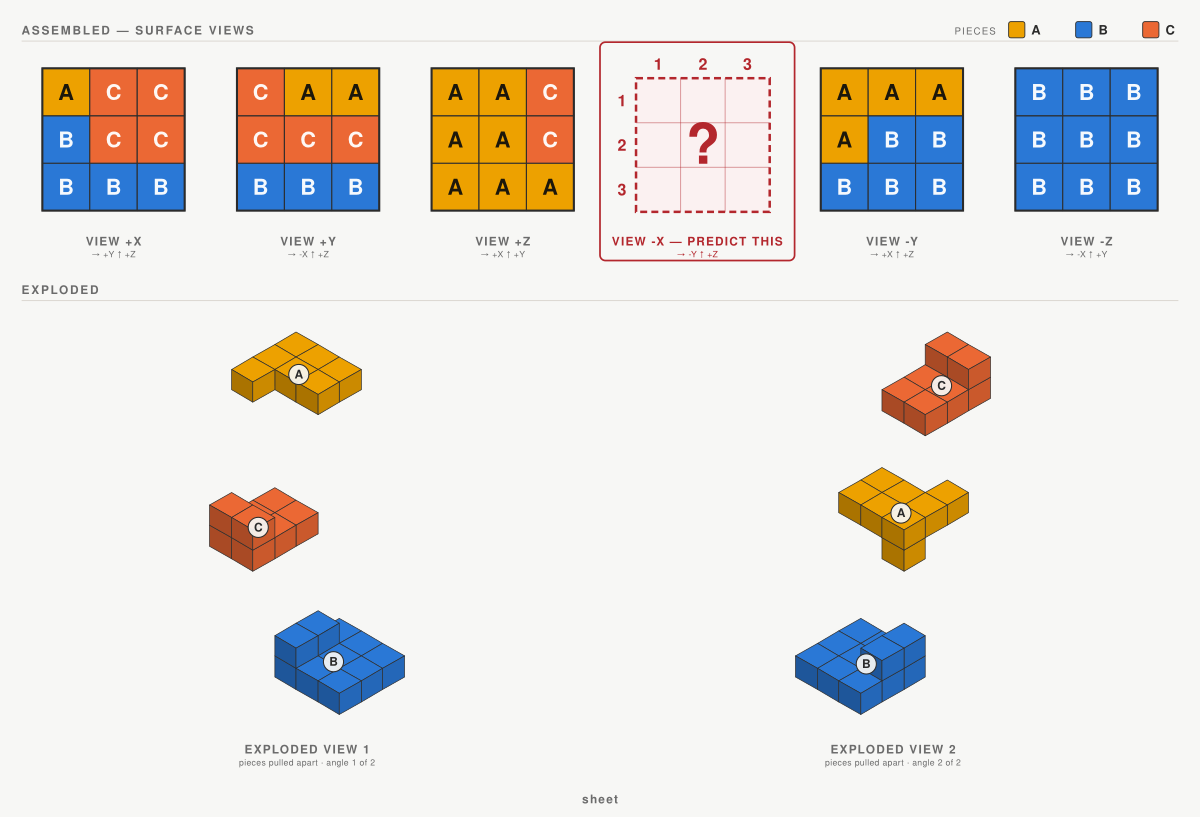}
    {\footnotesize (a) $3^3$ workspace with 3 pieces}
  \end{minipage}
  \hfill
  \begin{minipage}[t]{0.475\textwidth}
    \centering
    \includegraphics[width=\linewidth]{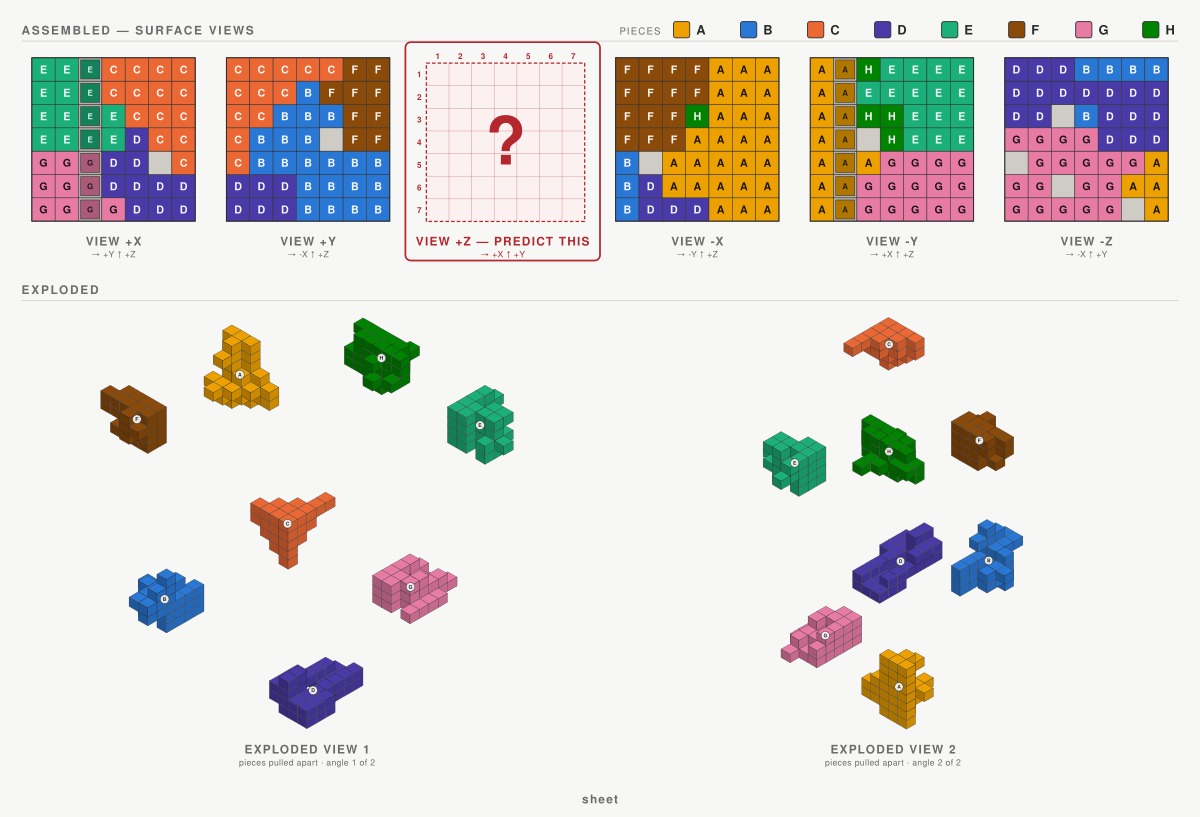}
    {\footnotesize (b) $7^3$ workspace with 8 pieces}
  \end{minipage}
  \caption{Benchmark overview and scale. Each item contains five observed
  projections, one held-out projection, and two separated views of every
  piece. The $3^3$ example shows the basic task. The $7^3$ example illustrates
  the denser occlusion and larger piece vocabulary at higher levels. Exact
  shapes and label grids are used for dataset validation; models receive one of the
  four representations defined in \cref{sec:experiments}.}
  \label{fig:overview}
\end{figure*}
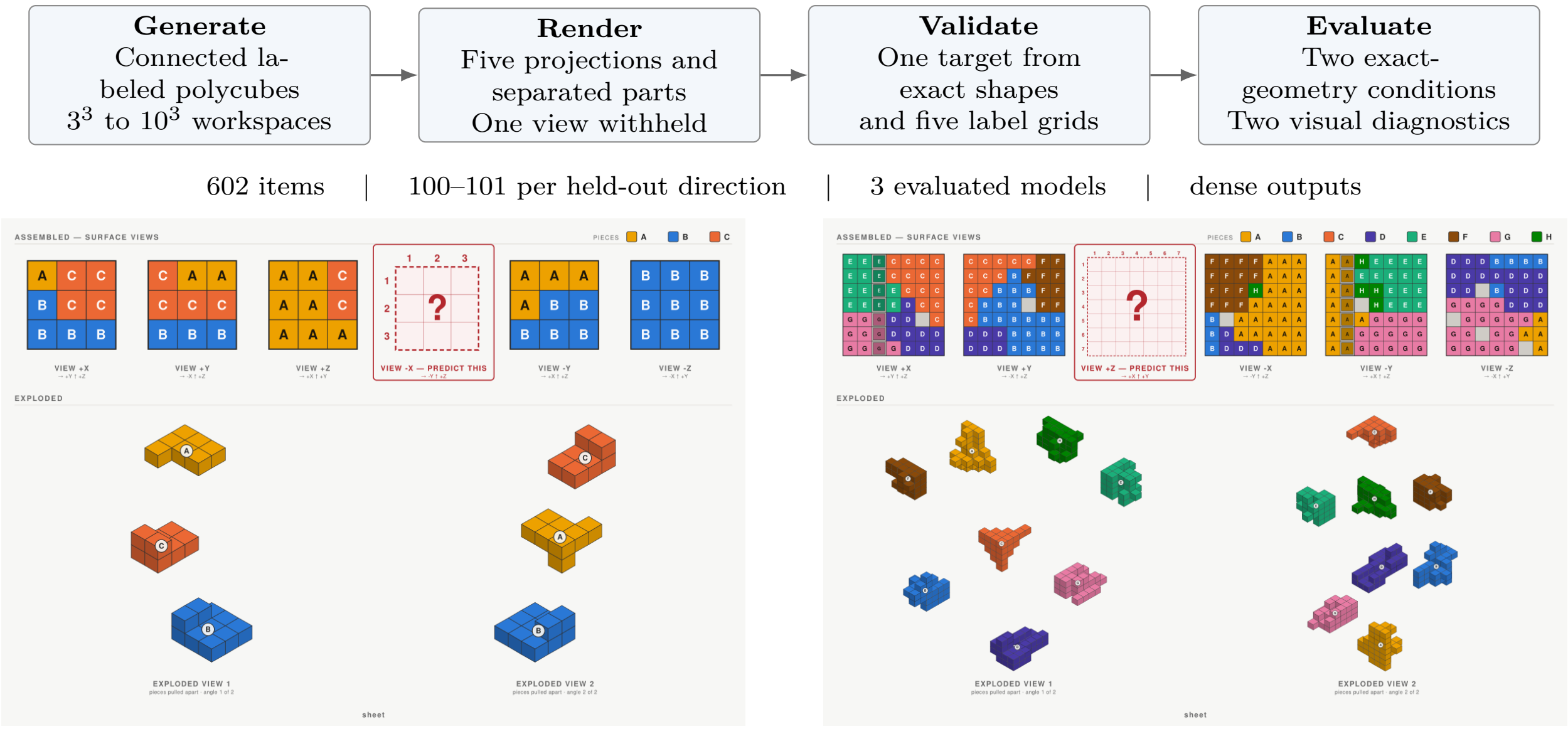

\subsection{Task definition}

Let $V:\{0,\ldots,n-1\}^3\rightarrow\{0,1,\ldots,k\}$ be a voxel
assembly. Label $0$ denotes empty space, and each nonzero label identifies one
connected piece. For a direction $d\in\{\pm X,\pm Y,\pm Z\}$, let
$R_d(i,j)=(x_1,\ldots,x_n)$ be the camera-to-object ray through image cell
$(i,j)$. For a nonempty ray, define
\begin{equation}
t_d(i,j)=\min\{s:V(x_s)\neq0\},\qquad F_d(i,j)=V(x_{t_d(i,j)}).
\label{eq:projection}
\end{equation}
If the ray is empty, we set $F_d(i,j)=\texttt{.}$. An item withholds one
direction $h$ and asks the model to generate $F_h$ as $n$ strings of length
$n$.

The abstract geometric evidence consists of the five grids $F_d$ for
$d\neq h$ and the exact labeled shape of every piece. Arms 1 and 3 expose this
evidence as label grids and local voxel coordinates and are the primary
conditions covered by the exact-geometry validation. Arms 2 and 4 instead show two static renders of
each separated piece. They are secondary visual diagnostics: performance also
depends on recovering shape from pixels, and the renders can leave piece
geometry ambiguous. Exact text supplies the piece shapes; placement, cross-view
alignment, occlusion, and the target-grid serialization remain to be resolved.
Piece position is always unknown. On 171
benchmark items, the separated pieces appear in rotated poses and their
assembled orientation must also be recovered.

Rendered surface cells distinguish a flush hit, a recessed hit, and an empty
ray. The scored grid records only the first-hit piece label or the empty symbol.
Every panel gives its screen-right and screen-up axes. This convention fixes
the orientation of the requested output without requiring a canonical front
view.

\subsection{Procedural instance construction}

Generation separates geometric diversity from answer validation. A proposal
begins with the complete $n^3$ lattice. Straight channels remove entire lines
of voxels and expose interior geometry from two opposing directions. Rare
enclosed cavities reduce visible evidence. Every removal preserves connectivity
of the occupied volume.

The remaining cells are partitioned into $k$ pieces by randomized multi-source
growth. Each seed expands through face-adjacent unassigned cells until the
volume is covered. We reject stalled growth, disconnected pieces, and pieces
with fewer than three voxels. The resulting labeled polycubes contain irregular
tabs, notches, and cross-view boundaries while still forming one connected
assembly.

Eight levels vary workspace size, piece count, channel frequency, and cavity
frequency. Sizes increase contiguously from $3^3$ to $10^3$, and piece counts
range from 2 to 10. At sizes $4^3$ and above, approximately one third of items
rotate every separated piece away from its assembled pose. Rotation is crossed
with scale rather than reserved for the largest instances.

\subsection{Acceptance criteria and corpus}

Every proposal must satisfy four conditions. First, the occupied volume and
each labeled piece are connected, nonempty, and in bounds. Second, none of the
five observed grids is identical to the target. Third, a per-item constant
label may score at most 60\%, and the mirrored opposite view may score at most
75\%. Fourth, the verifier from \cref{sec:certification} must establish that
the exact labeled shapes and five observed grids determine one target.

The benchmark contains 602 assemblies satisfying all four criteria, with
101 items each for $+X$ and $+Y$ and 100 for each remaining held-out direction.
\Cref{tab:data} summarizes the distribution by workspace size and piece count.

\begin{table}[t]
\centering
\small
\setlength{\tabcolsep}{4pt}
\begin{tabular}{@{}rrr@{}}
\toprule
$n$ & Items & Piece range \\
\midrule
3  & 58 & 2 to 3  \\
4  & 80 & 3 to 5  \\
5  & 88 & 4 to 7  \\
6  & 88 & 4 to 8  \\
7  & 80 & 5 to 9  \\
8  & 80 & 5 to 10 \\
9  & 64 & 6 to 10 \\
10 & 64 & 6 to 10 \\
\midrule
All & 602 & 2 to 10 \\
\bottomrule
\end{tabular}
\caption{Benchmark composition by workspace size.}
\label{tab:data}
\end{table}

\subsection{Representations and leakage control}

The renderer creates six ordered surface slots, including a placeholder at the
withheld direction. It also creates two separated views of the pieces and one
merged sheet. Letters and colors redundantly encode identity. The text
transcription gives the five letter grids, flush or recessed state, explicit
axis conventions, and each piece in local coordinates. Global placement is
absent.

The evaluation runner constructs an arm-specific payload containing only the
prompt and images assigned to that condition. It does not expose the dataset
manifest or answer files to the model. The dataset stores exact geometry for
auditing. Target renders and assembled views are stored separately from the
model-facing inputs. Every evaluation row stores a prompt hash, while the
dataset manifest hashes the
prompt file and fixes the ordered image filenames. A changed prompt is treated
as a new evaluation condition. We also reject any item whose
target matches a shown panel or closely copies a worked formatting example.

\subsection{Diagnostic subsets and shortcut baselines}

A target cell is \emph{directly recoverable} when its target ray is empty or
its first-hit voxel is also a first hit in at least one observed direction. The
opposite observed view reveals every empty target ray. The remaining cells form
the hard-cell set $\mathcal{H}$. This definition uses the ground-truth geometry only
to construct an evaluation mask; it never changes the input or target.
Across the 602 items, 14,677 of 27,794 target cells (52.8\%) are directly
recoverable and 13,117 (47.2\%) are hard; only 17 of 602 items contain no hard
cell.

Three shortcut references can calibrate cell accuracy: a per-item majority
label, the transformed opposite view, and an oracle that copies directly
recoverable cells before filling the remainder. The oracle uses target
information to select its fill label and is therefore a diagnostic stress test
rather than a usable solver. The majority and correctly target-oriented
opposite-view baselines obtain 43.5\% and 21.2\% mean per-item cell accuracy,
respectively, and neither solves any item exactly.

%% file: sec/4_certification.tex
\section{Dataset validation and target checks}
\label{sec:certification}

Orthographic observations can be compatible with several latent assemblies.
Target validation therefore requires every legal assembly consistent with an
item's exact evidence to yield the same held-out grid. The check is run
separately for every item and is bound to its five
first-hit label grids, exact labeled voxel shapes, held-out axis, target, and
content hash. The guarantee covers Arms 1 and 3, which expose exact piece
coordinates; shape identifiability from the static renders in Arms 2 and 4
remains unverified.

For each piece, the validator enumerates all 24 proper cube rotations and every
in-bounds integer translation. It removes placements that contradict an empty
observed ray, fail to intersect a ray showing that piece, or put another label
at the camera-nearest boundary voxel. Search then enumerates the remaining
placement combinations, pruning only overlaps. Every complete combination is
exactly reprojected onto the five observed faces. An item passes when the
planted assembly remains feasible and every evidence-consistent assembly
produces the stored held-out grid. Multiple latent assemblies are allowed when
their scored targets agree.

The check deliberately omits connectivity and mechanical-support constraints
on alternative assemblies, enlarging the candidate set.
It also omits rendered flush and recessed markers. The guarantee therefore
holds using a subset of the evidence supplied in the text-supported conditions.

Across the benchmark, 550 items have a singleton placement domain for every
piece and 52 require multiple candidates. The largest Cartesian candidate
product is 42 and the largest overlap-pruned search visits 31 nodes.

\begin{table}[t]
\centering
\scriptsize
\setlength{\tabcolsep}{4pt}
\begin{tabular}{@{}lr@{}}
\toprule
Validation property & Items \\
\midrule
Unique held-out target & 602 \\
One candidate placement per piece & 550 \\
Multiple candidates for at least one piece & 52 \\
\bottomrule
\end{tabular}
\caption{Exhaustive target validation of the 602-item benchmark.
Candidate counts refer to piece placements after filtering.}
\label{tab:certificate-audits}
\end{table}

Each JSONL certificate records the algorithm version, item SHA-256,
candidate counts, Cartesian search size, visited-node count, and reachable
target-face set. The verification program independently reconstructs candidates
and projections from the dataset. It also re-derives the 24 rotations and checks
its camera-ray convention against the generation projection. The dataset
manifest stores separate hashes for the dataset, shortcut records, validator,
and validation certificates, supporting verification of their consistency.
A full 602-item validation run takes 110 seconds (0.183
seconds/item, 103 MiB peak memory) on a 16-core Apple M4 Max.

%% file: sec/5_experiments.tex
\section{Experimental design}
\label{sec:experiments}

\paragraph{Research questions.}
The evaluation is organized around three paired questions. \textbf{RQ1} asks
how performance differs between exact textual geometry and rendered visual
evidence. \textbf{RQ2} asks whether adding exact symbolic evidence helps when
the images remain available. \textbf{RQ3} asks how strongly performance depends
on presenting the visual panels separately or on one sheet.

\subsection{Item-matched input conditions}

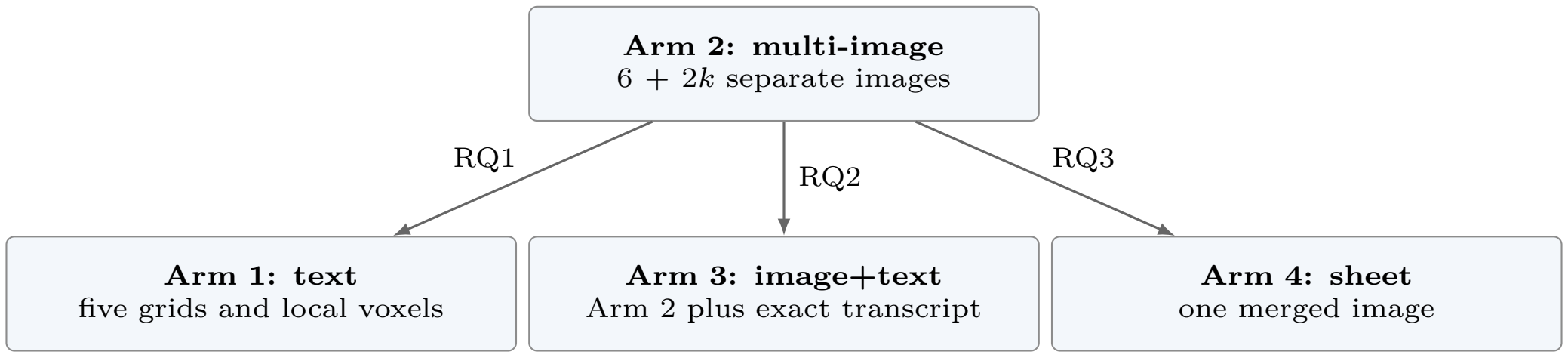
\begin{figure}[t]
\centering
\begin{tikzpicture}[
  arm/.style={draw=black!45, rounded corners=2pt, fill=wacvblue!6,
    align=center, text width=0.26\linewidth, minimum height=10mm,
    inner sep=2.5pt, font=\scriptsize},
  contrast/.style={-{Latex[length=1.5mm]}, semithick, draw=black!60,
    font=\scriptsize, align=center}
]
  \node[arm] (a2) {\textbf{Arm 2: multi-image}\\$6+2k$ separate images};
  \node[arm, below left=10mm and 1mm of a2] (a1) {\textbf{Arm 1: text}\\five grids and local voxels};
  \node[arm, below=10mm of a2] (a3) {\textbf{Arm 3: image+text}\\Arm 2 plus exact transcript};
  \node[arm, below right=10mm and 1mm of a2] (a4) {\textbf{Arm 4: sheet}\\one merged image};
  \draw[contrast] (a2) -- node[above left=-1mm] {RQ1} (a1);
  \draw[contrast] (a2) -- node[right] {RQ2} (a3);
  \draw[contrast] (a2) -- node[above right=-1mm] {RQ3} (a4);
\end{tikzpicture}
\caption{Paired experimental contrasts. RQ1 compares exact text with rendered
evidence. RQ2 adds exact symbolic evidence to the images. RQ3 jointly changes
image count and effective panel resolution by composing the same panels into
one sheet.}
\label{fig:arms}
\end{figure}

All conditions use the same 602 items and the same output contract. Arm 1 gives
the five label grids, their flush or recessed state, the axis convention, and
each piece as local voxel coordinates. Arm 2 supplies six ordered surface slots
and two isolated images per piece. Arm 3 adds the Arm 1 transcription to the Arm
2 images. Arm 4 contains the same visual panels as Arm 2 in one composite. The
task and output contract are shared; no condition receives the target grid or
global piece placement.

The contrasts in \cref{fig:arms} jointly vary information and representation.
Arm 1 provides exact piece geometry, while image-only inputs can leave it
ambiguous even for a perfect observer. Arm 3 adds both more precise geometric
information and a symbolic interface. Arm 4 changes
image count together with per-panel resolution. Arms 1 and 3 are primary
conditions covered by the exact-geometry validation domain. Arms 2 and 4 are
secondary visual diagnostics because their static renders can leave piece
geometry ambiguous.

\subsection{Models and repetitions}

We evaluate Gemini 3.7 Flash, GPT-5.5 (\texttt{gpt-5.5}), and Claude Opus 4.8.
The evaluation schedules three attempts for every model, condition, and benchmark
item: $3\times4\times602\times3=21{,}672$ evaluation keys, or 1,806 per
model-condition cell. Exact endpoint identifiers and inference controls are
retained for every call. An endpoint substitution is a separate condition and
is never pooled with the requested target.

Each endpoint uses a documented, provider-specific reasoning setting and output
limit. Model-to-model comparisons therefore characterize the deployed systems
under their respective inference settings.

The evaluation ledger contains one sanitized row per evaluation key, recording
the model and endpoint, arm, item, repetition, prompt hash, request and parse
status, final parsed grid, scores, effort and token metadata, latency,
and timestamp. The hashed dataset manifest fixes the item, prompt, answer,
shortcut, and certificate files. Credentials, internal URLs, raw replies, and
rationales remain private. Each model--condition--item--repetition combination
appears once, giving 21,672 scheduled keys: 21,608 observed records
and 64 explicit missing records, all from Opus. Our primary comparison uses the
common complete-case panel of 552 items for which every repetition is present
for every model and arm. This avoids treating concentrated Opus image-call
missingness as model error. We separately report the scheduled-run endpoint on
all 602 items, assigning zero to malformed, empty, truncated, or missing
attempts. Because archived logs cannot distinguish every unmade request from
an infrastructure failure, that secondary Opus endpoint is labeled incomplete.

\subsection{Metrics and statistical analysis}

Exact-grid accuracy is primary because it requires a complete reconstruction in
the requested frame. Cell accuracy gives partial credit. We average cell metrics
per item before aggregation, giving each item equal weight across grid sizes. Each
item's three attempts form one statistical cluster.

Across three trials, pass@3 marks an item solved at least once. Pass$^3$ marks
an item solved in all three attempts. The public ledger records parse status,
response coverage, token use, and latency for separate analysis. Malformed and
empty attempted outputs remain in the primary denominator.

Confidence intervals use a deterministic percentile bootstrap with 10,000
resamples of the 552 complete-case item-level means. Arm effects resample paired item
differences. Two-sided exact sign tests compare the direction of each nonzero
item-level difference, dropping ties; Holm correction covers the nine primary
model-by-question tests. Subgroup analyses are descriptive. These intervals
describe sensitivity to benchmark composition at the evaluated endpoint versions.

%% file: sec/6_results.tex
\section{Results}
\label{sec:results}

\subsection{Validated structured spatial reasoning}

\Cref{tab:main-results} reports the primary common complete-case analysis. Its
552 items have all three repetitions observed for every model and arm,
restricting comparisons to fully observed evaluations on a shared item set.
The full 602-item scheduled-run
endpoint, including 64 explicit missing Opus records, is reported separately
in the supplement and gives the same qualitative ordering.

\begin{table}[t]
\centering
\scriptsize
\setlength{\tabcolsep}{3.2pt}
\begin{tabular}{@{}lrrrr@{}}
\toprule
& \multicolumn{2}{c}{Validated exact geometry} & \multicolumn{2}{c}{Visual diagnostics} \\
\cmidrule(lr){2-3}\cmidrule(l){4-5}
Model & Text & Image+text & Images & Sheet \\
\midrule
Gemini 3.7 & 77.7 / 96.8 & \textbf{79.8} / 97.2 & 22.4 / 88.2 & 11.9 / 82.3 \\
GPT-5.5 & \textbf{33.6} / 85.5 & 33.2 / 84.6 & 13.5 / 75.2 & 6.8 / 59.6 \\
Opus 4.8 & \textbf{85.6} / 98.0 & 85.4 / 98.0 & 22.4 / 85.3 & 11.4 / 58.6 \\
\bottomrule
\end{tabular}
\caption{Primary exact-grid / cell accuracy (\%) on the common complete-case
panel ($N=552$), with three repetitions per item. The first two columns are
covered by the exact-geometry target check; the last two are secondary visual
diagnostics. Bold marks the higher exact accuracy within the primary conditions.}
\label{tab:main-results}
\end{table}

Adding images to exact text yields small changes in accuracy: image+text differs
from text by $+2.1$ points for Gemini (95\% CI $[0.0,4.2]$), $-0.5$ for GPT-5.5
($[-2.4,1.4]$), and $-0.2$ for Opus ($[-2.0,1.7]$). Exact accuracy in the
combined condition ranges from 33.2\% for GPT-5.5 to 85.4\% for Opus.
Exact text already supplies the shapes and observed
labels, so these scores primarily reflect symbolic assembly, coordinate handling,
and serialization.

\subsection{Representation, packaging, and reliability}

Adding exact text to separate images raises exact accuracy by 57.4 points for
Gemini (95\% CI $[53.9,60.9]$), 19.7 for GPT-5.5 ($[17.0,22.3]$), and 63.0 for
Opus ($[59.6,66.4]$). This gap reflects differences in geometric information
and representation, together with possible visual extraction errors. Ambiguity
in static piece renders further limits attribution of the gap to perception.
Separate panels outperform the merged
sheet by 10.5 points for Gemini ($[8.2,12.8]$), 6.7 for GPT-5.5
($[4.9,8.6]$), and 11.1 for Opus ($[8.7,13.5]$). This contrast jointly changes
image count and effective panel resolution.

For image+text, pass@3 / pass$^3$ is 90.9 / 67.9 for Gemini, 45.5 / 22.5 for
GPT-5.5, and 94.4 / 73.2 for Opus. The combined condition exclusively solves
25, 28, and 13 items, respectively, that neither text nor separate images
solves once; we call these \emph{combined-only solved} items. Across all three
models, 19/552 items (3.4\%) are never solved in image+text.

\paragraph{Scale and task difficulty.}
At $n=9$ and $n=10$, separate-image and sheet exact accuracy approach zero,
while exact-geometry conditions remain substantially stronger but decline with
size. Workspace size also changes piece count and occlusion, and rotated
presentations occur only at sizes $4^3$ and above. These covarying factors limit
interpretation of the descriptive size trends.

%% file: sec/7_failure.tex
\section{Limitations and broader impact}

\benchmark uses synthetic voxel geometry, cardinal cameras, and a small piece
alphabet. Generalization to natural-image reconstruction, continuous pose,
metric uncertainty, dynamics, or physical assembly requires further evaluation.
Exact coordinates deliberately
remove shape extraction from the primary conditions, which measure symbolic
assembly, coordinate transformation, and serialization. The automated target check
applies only when exact labeled shapes are supplied (Arms 1 and 3). Arms 2 and
4 use two static renders per piece, so their scores can mix visual ambiguity,
perception, assembly reasoning, and serialization. A separate audit of generated
piece renders finds a same-size alternative with the same two-render appearance
for 26.2\% of tested pieces. This piece-level lower bound leaves item-level target
ambiguity unresolved, since the assembled views may disambiguate individual pieces.
The ground-truth-visible three-rater audit in the supplement assesses rendering
quality; blinded human performance and pixel-level uniqueness remain open
evaluation questions.

The three endpoints use different reasoning controls and output limits, so the
scores characterize deployed systems under their respective inference settings.
Comparisons with open-weight VLMs and independent classical reconstruction
solvers remain for future work. The exhaustive validator checks exact-geometry
feasibility and target agreement against the known answer. The sanitized
evaluation ledger supports metric recomputation from prompts, hashes, parsed
final grids, statuses, and scores. Analysis of model-internal reasoning remains
outside its scope.

%% file: sec/8_conclusion.tex
\section{Conclusion}

\benchmark combines dense hidden-view prediction with exhaustive target
validation to diagnose structured 3D spatial reasoning and assembly. On the
primary 552-item panel, adding exact geometry to images gains 19.7--63.0
accuracy points, while adding images to exact text changes accuracy by only
$-0.5$ to $+2.1$ points. These findings characterize representation sensitivity
in a symbolic assembly task. The image-only variants remain uncertified, and
their scores may combine visual extraction errors, geometric ambiguity, and
assembly errors. The
benchmark's value is a reproducible, cell-level test of discrete spatial
operations with exact evidence; general visual 3D understanding requires
additional evaluation beyond this synthetic setting.

%% file: sec/supplement.tex
\section{Supplementary overview}

This supplement details the coordinate convention, input conditions, evaluation
protocol, and validation procedure for the 602-item benchmark.

\section{A benchmark example}
\label{sec:supp-example}

Item \texttt{mb3-000} is a $3\times3\times3$ workspace containing three
connected pieces. Its withheld direction is $-X$. Under the screen
convention, screen-right is $-Y$, screen-up is $+Z$, and the camera lies on the
$-X$ side and looks inward along $+X$. Thus right $\times$ up equals the camera
side direction: $(-Y)\times(+Z)=-X$.

\begin{table}[H]
\centering
\small
\begin{tabular}{@{}ll@{}}
\toprule
Field & Value \\
\midrule
Workspace & $3\times3\times3$ \\
Pieces & A, B, C \\
Piece sizes & 8, 11, 8 voxels \\
Withheld view & $-X$ \\
Screen orientation & right $-Y$, up $+Z$ \\
Target & \texttt{["AAA","CCA","BBB"]} \\
\bottomrule
\end{tabular}
\caption{Geometry and target for \texttt{mb3-000}.}
\label{tab:supp-example}
\end{table}

The five shown faces in the text condition are:

\begin{Verbatim}[fontsize=\scriptsize]
VIEW +X   (right +Y, up +Z)   VIEW +Y   (right -X, up +Z)
  ACC SSS                       CAA SSS
  BCC SSS                       CCC SSS
  BBB SSS                       BBB SSS

VIEW +Z   (right +X, up +Y)   VIEW -Y   (right +X, up +Z)
  AAC SSS                       AAA SSS
  AAC SSS                       ABB SSS
  AAA SSS                       BBB SSS

VIEW -Z   (right -X, up +Y)
  BBB SSS
  BBB SSS
  BBB SSS
\end{Verbatim}

The exact piece shapes are:

\begin{Verbatim}[fontsize=\scriptsize,breaklines=true]
Piece A: (0,0,0) (0,0,1) (0,1,1) (0,2,1)
         (1,0,1) (1,1,1) (1,2,1) (2,0,1)
Piece B: (0,0,0) (0,1,0) (0,2,0) (1,0,0)
         (1,0,1) (1,1,0) (1,2,0) (2,0,0)
         (2,0,1) (2,1,0) (2,2,0)
Piece C: (0,0,0) (0,1,0) (1,0,0) (1,1,0)
         (2,0,0) (2,0,1) (2,1,0) (2,1,1)
\end{Verbatim}

The withheld face, written top-to-bottom and left-to-right, is
\begin{Verbatim}[fontsize=\small]
["AAA", "CCA", "BBB"]
\end{Verbatim}

A correct Arm-1 response from the evaluation states that rows
correspond to $z=2,1,0$ and columns to $y=2,1,0$, and returns the target above.
The same model also solves Arms 2 and 3 on this item; its Arm-4 response is
\texttt{["AAA","CAA","BBB"]}, a one-cell error. Only the final grid is scored.

The validation certificate binds this item by SHA-256 and records one candidate
for each piece, three visited placement nodes, and the singleton face set
\texttt{[["AAA","CCA","BBB"]]}.
This example is included in the 602-item benchmark; every model run preceded
manuscript disclosure. Future hidden-test evaluations should exclude this
published example.

\section{The four input conditions}
\label{sec:supp-arms}

Every condition asks for the same withheld grid and uses the same JSON response
contract. Only the representation of geometric evidence changes.

\paragraph{Arm 1: text.}
The five observed grids include the first-hit labels, flush/recessed state, and
per-view screen axes. Every piece is listed as local voxel coordinates. This is
a primary condition in the exact-geometry certificate domain.

\paragraph{Arm 2: separate images.}
The model receives six ordered surface slots followed by two isolated renders
per piece. The two renders expose all six outer directions but need not reveal
interior occupancy uniquely. This arm is an uncertified visual diagnostic.

\paragraph{Arm 3: images with transcription.}
The model receives the complete Arm-2 image sequence plus the Arm-1 exact
transcription. The exact coordinates place this primary condition inside the
certificate domain; the images are redundant evidence.

\paragraph{Arm 4: merged sheet.}
All visual panels are composed into one image. This arm is an uncertified visual
diagnostic and jointly changes image count and effective panel resolution.

A separate audit of generated piece renders finds a same-size rearrangement
with the same two-render appearance for 26.2\% of tested pieces. This
piece-level lower bound leaves item-level target ambiguity unresolved: the five
observed faces may disambiguate individual pieces. Certifying targets from
rendered evidence requires an additional item-level analysis.

\paragraph{Rendered-input quality-control audit.}
The 48 audited items cross all eight workspace sizes with all six held-out
directions, one item per size--direction cell. Three raters independently see
the Arm-2 rendered evidence, requested direction and orientation convention,
and ground-truth grid. They answer four categorical questions: whether the
ground truth is consistent with the pieces, whether the orientation is clear,
whether symbols are readable, and whether another answer could reasonably fit.
All three answer yes to the first three and no to the fourth for every item.
Thus 48/48 items pass unanimously, with no reported error, ambiguity, missing
rating, or disagreement. Raw pairwise and three-way agreement is 100\%.
For each question separately, all ratings occupy one category, so the
chance-agreement denominator is zero and Cohen's and Fleiss' kappa are
undefined. This ground-truth-visible audit assesses rendering quality; blinded
human performance and pixel-level uniqueness require separate evaluation.

Across the 602 benchmark items, 14,677 of 27,794 target cells (52.8\%) are directly
recoverable from an empty target ray or a voxel exposed in another observed
view; 13,117 cells (47.2\%) are hard under this definition. Only 17 items have
no hard cells. The hard-cell fraction grows from 7.9\% at $n=3$ to 57.4\% at
$n=10$. A per-item majority-label baseline obtains 43.5\% mean cell accuracy,
and a correctly target-oriented opposite-view baseline obtains 21.2\%; neither
produces an exact answer on any of the 602 items.

\section{Evaluation estimands}
\label{sec:supp-estimands}

Let $y_{mair}\in\{0,1\}$ indicate an exact answer from model $m$, arm $a$,
item $i$, and repetition $r$. Let $c_{mair}\in[0,1]$ be cell accuracy, with
$c_{mair}=0$ for an attempted but unparseable response. For the common
complete-case panel, $N=552$ and $R=3$, and the primary estimators are
\begin{equation}
\widehat p_{ma}=\frac{1}{N}\sum_{i=1}^{N}\frac{1}{R}
\sum_{r=1}^{R}y_{mair},\qquad
\widehat c_{ma}=\frac{1}{N}\sum_{i=1}^{N}\frac{1}{R}
\sum_{r=1}^{R}c_{mair}.
\end{equation}

All 36 model--arm--repetition records are observed on these 552 items. A
secondary scheduled-run endpoint uses all 602 items and assigns zero to
malformed, empty, truncated, and missing scheduled responses. The ledger
preserves these statuses explicitly, allowing response coverage to be examined
alongside accuracy. Missing entries may correspond to unsent requests or
infrastructure failures; their cause remains unresolved in the archived logs.

For confidence intervals, define
$\bar y_{mai}=R^{-1}\sum_r y_{mair}$. We use a deterministic percentile
bootstrap with 10,000 resamples of the 552 item means. Arm contrasts resample
paired item differences. Two-sided exact sign tests drop ties, and Holm
correction covers the nine primary model-by-question comparisons.

For reliability,
$S_{ma}^{@3}=\{i:\max_r y_{mair}=1\}$ contains items solved at least once, and
$S_{ma}^{3}=\{i:\min_r y_{mair}=1\}$ contains items solved in all three
attempts. These sets require a status row for every scheduled key, including an
explicit zero-valued row when the scheduled-run response is missing.

\section{Additional evaluation results}
\label{sec:supp-results}

The three-model design contains 21,672 scheduled evaluation keys. The public
ledger has one row for every key: 21,608 observed responses and 64 explicit
missing Opus records (4 text, 54 images, 6 image+text, and 0 sheet). Scores
in the primary analysis use the common 552-item panel on which all 36
model--arm--repetition records are observed.

\begin{table}[H]
\centering
\scriptsize
\setlength{\tabcolsep}{2pt}
\begin{tabular}{@{}llrrrr@{}}
\toprule
Model & Metric & Text$^\dagger$ & Images & Image+text$^\dagger$ & Sheet \\
\midrule
Gemini 3.7 & Exact & 77.7 & 22.4 & \textbf{79.8} & 11.9 \\
 & Cell & 96.8 & 88.2 & 97.2 & 82.3 \\
GPT-5.5 & Exact & \textbf{33.6} & 13.5 & 33.2 & 6.8 \\
 & Cell & 85.5 & 75.2 & 84.6 & 59.6 \\
Opus 4.8 & Exact & \textbf{85.6} & 22.4 & 85.4 & 11.4 \\
 & Cell & 98.0 & 85.3 & 98.0 & 58.6 \\
\bottomrule
\end{tabular}
\caption{Primary common complete-case exact-grid and cell accuracy (\%) on
$N=552$. $^\dagger$Exact-geometry conditions.}
\label{tab:supp-exact}
\end{table}

\begin{table}[H]
\centering
\scriptsize
\setlength{\tabcolsep}{2pt}
\begin{tabular}{@{}lrrr@{}}
\toprule
Model & Text $-$ images & Image+text $-$ images & Images $-$ sheet \\
\midrule
Gemini 3.7 & 55.3 & 57.4 & 10.5 \\
GPT-5.5 & 20.2 & 19.7 & 6.7 \\
Opus 4.8 & 63.2 & 63.0 & 11.1 \\
\bottomrule
\end{tabular}
\caption{Primary paired-condition exact-accuracy differences (percentage
points) on the common complete-case panel.}
\label{tab:supp-contrasts}
\end{table}

\begin{table}[H]
\centering
\scriptsize
\setlength{\tabcolsep}{2pt}
\begin{tabular}{@{}lrrrr@{}}
\toprule
Model & Text & Images & Image+text & Sheet \\
\midrule
Gemini 3.7 & 78.2 / 96.8 & 21.0 / 88.2 & 80.2 / 97.1 & 11.0 / 82.1 \\
GPT-5.5 & 32.8 / 85.4 & 12.3 / 75.2 & 32.4 / 84.6 & 6.2 / 59.7 \\
Opus 4.8 & 85.5 / 97.7 & 20.8 / 82.8 & 85.4 / 97.8 & 10.4 / 59.5 \\
\bottomrule
\end{tabular}
\caption{Secondary scheduled-run exact / cell accuracy (\%) on $N=602$. Each
model--arm denominator is 1,806; 64 missing Opus records receive zero.}
\label{tab:supp-scheduled}
\end{table}

On the common panel, image+text pass@3 / pass$^3$ is 90.9/67.9 for Gemini,
45.5/22.5 for GPT-5.5, and 94.4/73.2 for Opus. Their combined-only solved
counts are 25, 28, and 13; 19 items are never solved in image+text by any
included model. The main paper reports paired confidence intervals for the
representation contrasts.

\section{Validation and reproducibility}

For exact piece geometry, the validator enumerates proper rotations and
in-bounds translations. Candidate filters use empty rays, required label--ray
intersections, and the nearest boundary voxel. These filters remove only
impossible placements. The validator then enumerates every remaining placement
combination, prunes overlaps, exactly reprojects the five observed faces, and
requires every surviving assembly to produce the same withheld target. It also
requires the planted assembly to survive the filters and reproduce the stored
answer, establishing existence.

Each JSONL certificate stores an algorithm version, item SHA-256, identity
fields, candidate counts, Cartesian search size, visited-node count, the set of
reachable target faces, and status flags. The verification program independently
reconstructs candidates and projections from the dataset.

The candidate filters preserve legal assemblies, exact reprojection removes
tuples that disagree with the supplied evidence, and the planted-assembly
check keeps the consistent set nonempty. Together, these checks establish
target uniqueness under the exact-geometry conditions.

All 602 benchmark items pass the target-uniqueness check. Of these, 550 items
have singleton candidate domains and 52 require multi-candidate search.
The largest Cartesian product is 42 and the
largest overlap-pruned audit visits 31 nodes. Every record stores the item hash,
identity fields, candidate counts, search size, node count, and singleton target
face set.

The self-contained verification program reproduces the held-out
target for all 602 items. On a 16-core Apple M4 Max with 64 GiB
RAM, one warm-cache end-to-end run takes 110 seconds (0.183 seconds/item) with
103 MiB peak memory.

\section{Evaluation records and data access}

Scoring uses each model's final parsed grid. Raw replies and free-text
rationales remain private. The public evaluation ledger contains one row per
evaluation key with the model and endpoint,
arm, item, repetition, prompt hash, request status, stop reason, final parsed
grid, parse status, exact and cell scores, effort and token metadata, latency,
and timestamp.

The reproducibility materials include the 602-item manifest, exact prompts,
image filename lists, answers, validation certificates and verification code,
shortcut records, the sanitized ledger, and table-generation outputs. The
model-facing image archive contains the input panels, component renders, and
merged sheets. File-level SHA-256 hashes support verification of the dataset
and evaluation artifacts.

\paragraph{Asset provenance and distribution.}
The benchmark's voxel assemblies, component shapes, and rendered task panels
are generated procedurally for this study. The accompanying repository
distributes benchmark data, rendered images, documentation, and validation code
under the Creative Commons Attribution-NonCommercial 4.0 license (CC BY-NC 4.0).
Model access remains subject to each provider's terms.

%% file: sec/supp_references.tex
\renewcommand{\refname}{Supplementary References}